\documentclass{article}

\usepackage[final]{nips_2016}

\usepackage[utf8]{inputenc} 
\usepackage[T1]{fontenc}    
\usepackage{hyperref}       
\usepackage{url}            
\usepackage{booktabs}       
\usepackage{amsfonts}       
\usepackage{nicefrac}       
\usepackage{microtype}      
\usepackage{algorithm}
\usepackage{algorithmic}
\usepackage{graphicx}

\usepackage{tabulary}
\usepackage{booktabs}

\usepackage{epsf,psfrag,amsfonts, xcolor,array}
\usepackage{amsmath}
\usepackage{multirow}
\usepackage{cases}
\usepackage{verbatim}
\usepackage{xspace}
\usepackage{bbm}
\usepackage{mathrsfs}
\usepackage{setspace}
\usepackage[fleqn,tbtags]{mathtools}
\usepackage{tikz}

\usetikzlibrary{positioning}

\usepackage[colorinlistoftodos, textwidth=20mm]{todonotes}
\usepackage{color}

\newcommand{\vecy}{\vec y}

\newcommand{\smucrl}{\textsc{\small{SM-UCRL}}\xspace}
\newcommand{\ucrl}{\textsc{\small{UCRL}}\xspace}

\renewcommand{\Re}{\mathbb{R}}
\newcommand{\calP}{\mathcal P}

\newcommand{\A}{\mathcal A}
\newcommand{\B}{\mathcal B}
\newcommand{\X}{\mathcal X}
\newcommand{\T}{\mathcal T}

\newcommand{\TN}{N}

\newcommand{\Y}{\mathcal Y}

\newcommand{\M}{\mathcal M}

\newcommand{\Prob}{\mathbb P}

\newcommand{\R}{\mathcal{R}}

\newcommand{\wt}{\widetilde}
\newcommand{\wh}{\widehat}
\newcommand{\wb}{\overline}

\newtheorem{theorem*}{Theorem}
\newtheorem{theorem}{Theorem}

\title{Experimental results : Reinforcement Learning of POMDPs using Spectral Methods}

\author{
  Kamyar Azizzadenesheli \\
  University of California, Irvine \\
  \And
  Alessandro Lazaric \\
  INRIA, France \\
  \And
  Animashree Anandkumar\\
  University of California, Irvine \\
}

\begin{document}

\maketitle
\begin{abstract}
We propose a new reinforcement learning algorithm for partially observable Markov decision processes (POMDP) based on spectral decomposition methods. While spectral methods have been previously employed for consistent learning of (passive) latent variable models such as hidden Markov models, POMDPs are more challenging  since the learner interacts with the environment and possibly changes the future observations in the process. We devise a learning algorithm running through epochs, in each epoch we employ spectral techniques to learn the POMDP parameters from a trajectory generated by a fixed policy. At the end of the epoch, an optimization oracle returns the optimal memoryless planning policy which maximizes the expected reward based on the estimated POMDP model. We prove an order-optimal regret bound with respect to the optimal memoryless policy and efficient scaling with respect to the dimensionality of observation and action spaces. 
\end{abstract}




\section{Introduction}\label{s:intro}

Reinforcement Learning (RL) is an effective approach to solve the problem of sequential decision--making under uncertainty. RL agents learn how to maximize long-term reward using the experience obtained by direct interaction with a stochastic environment~\citep{sutton1998introduction}. Since the environment is initially unknown, the agent has to balance between \textit{exploring} the environment to estimate its structure, and \textit{exploiting} the estimates to compute a policy that maximizes the long-term reward. As a result, designing a RL algorithm requires three different elements: \textbf{1)} an estimator for the environment's structure, \textbf{2)} a planning algorithm to compute the optimal policy of the estimated environment~\citep{lavalle2006planning}, and \textbf{3)} a strategy to make a trade off between exploration and exploitation to minimize the \textit{regret}, i.e., the difference between the performance of the exact optimal policy and the rewards accumulated by the agent over time.

Most of RL literature assumes that the environment can be modeled as a Markov decision process (MDP), with a Markovian state evolution that  is fully observed. A number of exploration--exploitation strategies have been shown to have strong performance guarantees for MDPs, either in terms of regret or sample complexity (see Sect.~\ref{ss:related} for a review). However, the assumption of full observability of the state evolution is often violated in practice, and the agent may only have noisy observations of the true state of the environment (e.g., noisy sensors in robotics). In this case, it is more appropriate to use the partially-observable MDP or POMDP~\citep{sondik1971the-optimal} model.

Many challenges arise in designing RL algorithms for POMDPs. Unlike in MDPs, the estimation problem (element 1) involves identifying the parameters of a latent variable model (LVM). In an MDP  the agent directly observes (stochastic) state transitions, and the estimation of the generative model is straightforward via empirical  estimators. On the other hand, in  a POMDP the transition and reward models must be inferred from noisy observations and the Markovian state evolution is hidden. The planning problem (element 2), i.e., computing the optimal policy for a POMDP with known parameters, is PSPACE-complete~\citep{papadimitriou1987the-complexity}, and it requires solving an augmented MDP built on a continuous belief space (i.e., a distribution over the hidden state of the POMDP). Finally, integrating estimation and planning in an exploration--exploitation strategy (element 3) with guarantees is non-trivial and no no-regret strategies are currently known (see Sect.~\ref{ss:related}). To handle these challenges, we build up the results in this paper on the top of the previous paper \cite{azizzadenesheli2016reinforcement} on RL of POMDPs.


\subsection{Summary of Results}\label{ss:summary}

The main contributions of this paper are as follows: We propose a new RL algorithm for POMDPs that incorporates spectral parameter estimation within a exploration-exploitation framework. Then we apply this algorithm on a grid world atari game and compare its performance with state of the art Deep Q Learning (DQN) \cite{mnih2013playing}. We show that when the underlying model is not $MDP$, model based $MDP$ algorithms learn wrong model representation and model free algorithms, learn Q-function on observation set which does not carry the markovian property anymore. Furthermore, because of non-markovianity on observation set, current observation is not sufficient statistic for policy and sort of memory is required. Assume the game which has one green and one red apples. At the beginning of the game, the emulator reveals a flag which shows which apple has positive reward and which one has negative reward. In this case, the MDP based learner forgets the flag and always suffers regret linear regret. \\
In this paper, the estimation of the POMDP is carried out via spectral methods which involve decomposition of certain moment tensors computed from data. This learning algorithm is interleaved  with  the optimization of the planning policy  using an exploration--exploitation strategy inspired by the \ucrl method for MDPs~\citep{jaksch2010near-optimal}. The resulting algorithm, called \smucrl (\textit{Spectral Method for Upper-Confidence Reinforcement Learning}), runs through epochs of variable length, where the agent follows a fixed policy until enough data are collected and then it updates the current policy according to the estimates of the POMDP parameters and their accuracy. \\
We derive a regret bound with respect to the best memoryless (stochastic) policy for the given POMDP. Indeed, for a general POMDP, the optimal policy need not be memoryless. However, finding the optimal policy is uncomputable for infinite horizon regret minimization~\citep{madani1998computability}. Instead memoryless policies have shown good performance in practice (see the Section on related work). Moreover, for the class of so-called {\em contextual MDP}, a special class of POMDPs, the optimal policy is also  memoryless~\citep{krishnamurthy2016contextual-mdps}.
\subsection{Related Work}\label{ss:related}
In last few decades, MDP has been widely studied ~\citep{kearns2002near,brafman2003r,jaksch2010near-optimal} in different setting. While RL in MDPs has been widely studied, the design of effective exploration--exploration strategies in POMDPs is still relatively unexplored. \citet{ross2007bayes} and \citet{poupart2008model-based} propose to integrate the problem of estimating the belief state into a model-based Bayesian RL approach, where a distribution over possible MDPs is updated over time. An alternative to model-based approaches is to adapt model-free algorithms, such as Q-learning, to the case of POMDPs. \citet{perkins2002reinforcement} proposes a Monte-Carlo approach to action-value estimation and it shows convergence to locally optimal memoryless policies. An alternative approach to solve POMDPs is to use policy search methods, which avoid estimating value functions and directly optimize the performance by searching in a given policy space, which usually contains memoryless policies (see e.g.,~\citep{ng2000pegasus}). Beside its practical success in offline problems, policy search has been successfully integrated with efficient exploration--exploitation techniques and shown to achieve small regret~\citep{gheshlaghi-azar2013regret}. Nonetheless, the performance of such methods is severely constrained by the choice of the policy space, which may not contain policies with good performance. \\
Matrix decomposition methods have been previously used in the more general setting of predictive state representation (PSRs)~\citep{boots2011closing} to reconstruct the structure of the dynamical system. Despite the generality of PSRs, the proposed model relies on strong assumptions on the dynamics of the system and it does not have any theoretical guarantee about its performance. Recently, \citep{hamilton2014efficient} introduced compressed PSR (CPSR) method to reduce the computation cost in PSR by exploiting the advantages in dimensionality reduction, incremental matrix decomposition, and compressed sensing. In this work, we take these ideas further by considering more powerful tensor decomposition techniques. \\
In last few decades, latent variable models have become popular model for the problems with partially observable variables. Traditional methods such as Expectation-Maximization (EM) and variational methods have been used to learn the hidden structure of the model but usually they have no consistency guarantees, they are computationally massive, and mostly converge to local optimum which can be arbitrarily bad. To over come these drawbacks, spectral methods have been used for consistent estimation of a wide class of LVMs \cite{anandkumar2012method}, \cite{anandkumar2014tensor}, the theoretical guarantee and computation complexity by using robust tensor power method are well studied in \cite{song2013nonparametric} and \cite{wang2015fast}. Today, spectral methods and tensor decomposition methods are well known as a credible alternative for EM and variational methods for inferring the latent structure of the model. These method have been shown to be efficient in learning of Gaussian mixture models, topic modeling, Latent Dirichlet Allocation, Hidden markov model, etc.




\section{Preliminaries}\label{s:preliminaries}

\begin{figure}[t]
\begin{center}
\begin{psfrags}
\psfrag{x1}[][1]{$x_t$}
\psfrag{x2}[][1]{$x_{t+1}$}
\psfrag{x3}[][1]{$x_{t+2}$}
\psfrag{y1}[][1]{$\vec{y}_t$}
\psfrag{y2}[][1]{$\vec{y}_{t+1}$}
\psfrag{r1}[][1]{${r}_t$}
\psfrag{r2}[][1]{${r}_{t+1}$}
\psfrag{a1}[][1]{$a_{t}$}
\psfrag{a2}[][1]{$a_{t+1}$}
\includegraphics[scale=0.3]{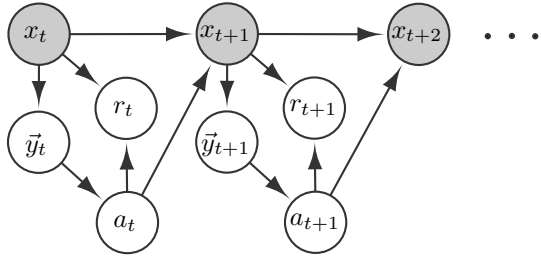}
\end{psfrags}
\end{center}
\vspace{-0.0in}
\caption{Graphical model of a POMDP under memoryless policies.}
\label{fig:pds2}
\end{figure}

A POMDP $M$ is a tuple $\langle \X, \A, \Y, \R, f_T, f_R, f_O\rangle$, where $\X$ is a finite state space $x$ with cardinality $|\X|=X$, $\A$ is a finite action space $a$ with cardinality $|\A|=A$, $\Y$ is a finite observation space $\vecy$ with cardinality $|\Y|=Y$ (the vector representation is w.r.t one-hot encoding), and $\R$ is a finite reward space $r$ with cardinality $|\R|=R$ and largest reward $r_{\max}$. Finally, $f_T$ denotes the transition density, so that $f_T(x'|x,a)$ is the probability of transition to $x'$ given the state-action pair $(x,a)$, $\bar{r}(x,a)$ is  mean reward at state $x$ and action $a$. Furthermore, $f_O$ is the observation density, so that $f_O(\vec{y}|x)$ is the probability of receiving the observation in $\Y$ corresponding to the indicator vector $\vec{y}$ given the state $x$. Whenever convenient, we use tensor forms for the density functions such that
\begin{align*}
T_{i,j,l} &= \Prob[x_{t+1}=j | x_t = i, a_t = l] = f_T(j|i,l)~,~ O_{n,i}= \Prob[\vec{y}=\vec{e}_n | x = i] = f_O(\vec{e}_n|i)
\end{align*}
Such that $T\in\Re^{X\times X\times A}$ and $O\in\Re^{Y\times X}$. We also denote by $T_{:, j, l}$ the fiber (vector) in $\Re^{X}$ obtained by fixing the arrival state $j$ and action $l$ and by $T_{:,:,l}\in\Re^{X\times X}$ the transition matrix between states when using action $l$. The graphical model associated to the POMDP is illustrated in Fig.~\ref{fig:pds2}. 

A policy is stochastic mapping from observations to actions and for any policy $\pi$ we denote by $f_\pi(a|\vec{y})$ its density function. We denote by $\calP$ the set of all stochastic memoryless policies. Acting according to a policy $\pi$ in a POMDP $M$ defines a Markov chain characterized by a transition density
\begin{align*}
f_{T,\pi}(x'|x) = \sum_a \sum_{\vec{y}} f_\pi(a|\vec{y}) f_O(\vec{y}|x) f_T(x'|x,a),
\end{align*}
and a stationary distribution $\omega_\pi$ over states such that $\omega_\pi(x) = \sum_{x'} f_{T,\pi}(x'|x)\omega_\pi(x')$. The expected average reward performance of a policy $\pi$ is
\begin{align*}
\eta(\pi; M) = \sum_x \omega_\pi(x) \wb{r}_\pi(x),
\end{align*}
where $\wb{r}_\pi(x)$ is the expected reward of executing policy $\pi$ in state $x$ defined as
\begin{align*}
\wb{r}_\pi(x) = \sum_a \sum_{\vec{y}} f_O(\vec{y}|x) f_\pi(a|\vec{y}) \wb{r}(x,a),
\end{align*}
The best stochastic memoryless policy in $\calP$ is $\pi^+ = \displaystyle\arg\max_{\pi\in\calP} \eta(\pi; M)$ and we denote by $\eta^+ = \eta(\pi^+; M)$ its average reward.\footnote{We use $\pi^+$ rather than $\pi^*$ to recall the fact that we restrict the attention to $\calP$ and the actual optimal policy for a POMDP in general should be constructed on the belief-MDP.}




\section{Learning the Parameters of the POMDP}\label{s:learning.pomdp}

In this section we introduce a novel spectral method to estimate the POMDP parameters $f_T$, $f_O$, and $f_R$. A stochastic policy $\pi$ is used to generate a trajectory $(\vec{y}_1, a_1, {r}_1, \ldots, \vec{y}_N, a_N, {r}_N)$ of $\TN$ steps. Similar to the case of HMMs, the key element to apply the spectral methods is to construct a multi-view model for the hidden states. Despite its similarity, the spectral method developed for HMM by~\cite{anandkumar2014tensor} cannot be directly employed here. In fact, in HMMs the state transition and the observations only depend on the current state. On the other hand, in POMDPs the probability of a transition to state $x'$ not only depends on $x$, but also on action $a$. Since the action is chosen according to a memoryless policy $\pi$ based on the current observation, this creates an indirect dependency of $x'$ on observation $\vec{y}$, which makes the model more intricate.


\subsection{The multi-view model}\label{ss:multi-view}

We estimate POMDP parameters for each action $l\in [A]$ separately. Let $t\in [2,N-1]$ be a step at which $a_t=l$, we construct three views $(a_{t-1}, \vec{y}_{t-1}, {r}_{t-1})$, $(\vec{y}_t, {r}_t)$, and $(\vec{y}_{t+1})$ which all contain observable elements. As it can be seen in Fig.~\ref{fig:pds2}, all three views provide some information about the hidden state $x_t$ (e.g., the observation $\vec{y}_{t-1}$ triggers the action $a_{t-1}$, which influence the transition to $x_t$). A careful analysis of the graph of dependencies shows that conditionally on $x_t, a_t$ all the views are independent. For instance, let us consider $\vec{y}_{t}$ and $\vec{y}_{t+1}$. These two random variables are clearly dependent since $\vec{y}_t$ influences action $a_t$, which triggers a transition to $x_{t+1}$ that emits an observation $\vec{y}_{t+1}$. Nonetheless, it is sufficient to condition on the action $a_t = l$ to break the dependency and make $\vec{y}_t$ and $\vec{y}_{t+1}$ independent. Similar arguments hold for all the other elements in the views, which can be used to recover the latent variable $x_t$. More formally, we encode the triple $(a_{t-1},\vec{y}_{t-1},{r}_{t-1})$ into a vector $\vec{v}_{1,t}^{(l)}\in \Re^{A\cdot Y}$, so that view $\vec{v}_{1,t}^{(l)} = \vec{e}_s$ whenever $a_{t-1}=k$, $\vec{y}_{t-1}=\vec{e}_n$, and ${r}_{t-1}=\vec{e}_m$ for a suitable mapping between the index $s\in\{1,\ldots,A\cdot Y\}$ and the indices $(k,n)$ of the action, observation, and reward. Similarly, we proceed for $\vec{v}_{2,t}^{(l)}\in\Re^{Y}$ and  $\vec{v}_{3,t}^{(l)}\in\Re^{Y}$. We introduce the three view matrices $V^{(l)}_\nu$ with $\nu\in\{1,2,3\}$ associated with action $l$ defined as $V^{(l)}_1\in\Re^{A\cdot Y\times X}$, $V^{(l)}_2\in\Re^{Y\cdot R\times X}$, and $V^{(l)}_3\in\Re^{Y\times X}$ such that
\begin{align*}
[V_1^{(l)}]_{s,i}&=\Prob\big(\vec{v}_1^{(l)}=\vec{e}_s | x_2=i\big) = [V_1^{(l)}]_{(n,m,k),i} = \Prob\big(\vecy_1=\vec{e}_n,=\vec{e}_m,a_1=k|x_2=i\big),\\[0.05in]
[V_2^{(l)}]_{s,i}&=\Prob\big(\vec{v}_2^{(l)}=\vec{e}_s |x_2=i,a_2=l\big) = [V_2^{(l)}]_{(n',m'),i} = \Prob\big(\vecy_2=\vec{e}_{n'}=\vec{e}_{m'}|x_2=i,a_2=l\big),\\[0.05in]
[V_3^{(l)}]_{s,i}&=\Prob\big(\vec{v}_3^{(l)}=\vec{e}_s |x_2=i,a_2=l\big) = [V_3^{(l)}]_{n'',i} = \Prob\big(\vecy_3=\vec{e}_{n''}|x_2=i,a_2=l\big).
\end{align*}
At the end, let's concatinat the reward of time $t-1$ and $t$ to the end of vectors $\vec{v}_{1,t}^{(l)}$ and $\vec{v}_{2,t}^{(l)}$ which ends up to one extra row at the bottom of $V_1^{(l)}$ and $V_2^{(l)}$. By simple manipulation, one can efficiently extract the model parameters out of $V_2^{(l)}$, and $V_3^{(l)}$ for $\forall l\in \A$
\begin{algorithm}[t]
\setstretch{1.1}
\begin{small}
\begin{algorithmic}
\STATE \textbf{Input:}
\STATE $\quad$ Policy density $f_\pi$, number of states $X$
\STATE $\quad$ Trajectory $\langle (\vec{y}_1,a_1,r_1), (\vec{y}_2,a_2,{r}_2),\ldots, (\vec{y}_N,a_N, {r}_N)\rangle$

\STATE \textbf{Variables:}
\STATE $\quad$ Estimated second and third views $\wh{V}_2^{(l)}$, and $\wh{V}_3^{(l)}$ for any action $l\in[A]$
\STATE $\quad$ Estimated observation, reward, and transition models $\wh{f}_O$, $\bar{r}$, $\wh{f}_T$
\STATE
\FOR{$l=1,\ldots,A$}
\STATE Set $\T(l) = \{t \in [N-1]: a_t = l\}$ and $N(l) = |\T(l)|$
\STATE Construct views $\enspace\vec{v}_{1,t}^{(l)} = (a_{t-1},\vec{y}_{t-1},{r}_{t-1}),$ $\enspace$ $\vec{v}_{2,t}^{(l)} = (\vec{y}_{t},{r}_{t}),$ $\enspace$ $\vec{v}_{3,t}^{(l)} = \vec{y}_{t+1}\quad$ for any $t\in\T(l)$
\STATE Symmetrize the views
\STATE Compute second and third moments
$$
\wh{M}_2^{(l)} = \frac{1}{N(l)} \sum_{t\in\T_l} \wt{\vec{v}}_{1,t}^{(l)} \otimes \wt{\vec{v}}_{2,t}^{(l)}, \quad
\wh{M}_3^{(l)} = \frac{1}{N(l)} \sum_{t\in\T_l} \wt{\vec{v}}_{1,t}^{(l)} \otimes \wt{\vec{v}}_{2,t}^{(l)} \otimes \vec{v}_{3,t}^{(l)}
$$
\STATE Compute $\wh{V}_3^{(l)}$ = \textsc{\small TensorDecomposition}($\wh{M}_2^{(l)}$, $\wh{M}_3^{(l)}$)
\ENDFOR
\STATE Estimate $\wh{f}_O$, $\bar{r}$, $\wh{f}_T$
\STATE \textbf{Return:} $\bar{r}$, $\wh{f}_T$, $\wh{f}_O$, $\B_R$, $\B_T$, $\B_O$
\end{algorithmic}
\end{small}
\caption{Estimation of the POMDP parameters. The routine \textsc{\small TensorDecomposition} refers to the spectral tensor decomposition method of~\cite{azizzadenesheli2016reinforcement}.}
\label{alg:spectral.pomdp}
\end{algorithm}

\paragraph{Empirical estimates of POMDP parameters.}
In practice, $M_2^{(l)}$ and $M_3^{(l)}$ are not available and need to be estimated from samples. Given a trajectory of $N$ steps obtained executing policy $\pi$, let $\T(l) = \{t\in[2,N-1]: a_t = l\}$ be the set of steps when action $l$ is played, then we collect all the triples $(a_{t-1}, \vec{y}_{t-1}, {r}_{t-1})$, $(\vec{y}_t, {r}_t)$ and $(\vec{y}_{t+1})$ for any $t\in\T(l)$ and construct the corresponding views $\vec{v}_{1,t}^{(l)}$, $\vec{v}_{2,t}^{(l)}$, $\vec{v}_{3,t}^{(l)}$. Then we symmetrize the views. Given the resulting $\wh{M}_2^{(l)}$ and $\wh{M}_3^{(l)}$, we apply the spectral tensor decomposition method to recover empirical estimates of second and third views $\wh{V}_2^{(l)}$, $\wh{V}_3^{(l)}$. Thereafter, a simple manipulation results in the model parameters.
 The overall method is summarized in Alg.~\ref{alg:spectral.pomdp}. The empirical estimates of the POMDP parameters enjoy the following guarantee.

\begin{theorem}[Learning Parameters]\label{thm:estimates}
Let $\wh{f}_O$, $\wh{f}_T$, and $\wh{f}_R$ be the estimated POMDP models using a trajectory of $N$ steps. we have
%
%
\begin{align}\label{eq:obs.bound}
\|\wh{f}_O(\cdot|i) \!-\! f_O(\cdot|i)\|_1 \leq \B_O^{(l)} \!:=C_O \sqrt{\frac{Y\log(1/\delta)}{N(l)}}~, ~\|\wh{f}_R(\cdot|i,l) - f_R(\cdot|i,l)\|_1 \leq \B_R^{(l)} := C_R \sqrt{\frac{\log(1/\delta)}{N(l)}},
\end{align}
\begin{align}\label{eq:transition.bound}
\| \wh{f}_T(\cdot|i,l) \!-\! f_T(\cdot|i,l) \|_2 \leq \B_T^{(l)} := C_T  \sqrt{\frac{YX^2\log(1/\delta)}{N(l)}},
\end{align}
%
%
\noindent with probability $1-6(Y^2+AY)A\delta$ (w.r.t. the randomness in the transitions, observations, and policy), where $C_O$, $C_R$, and $C_T$ are numerical constants.
\end{theorem}

\noindent\paragraph{Remark 1 (consistency and dimensionality).}
All previous errors decrease with a rate $\wt{O}(1/\sqrt{N(l)})$, showing the consistency of the spectral method, so that if all the actions are repeatedly tried over time, the estimates converge to the true parameters of the POMDP. This is in contrast with EM-based methods which typically get stuck in local maxima and return biased estimators, thus preventing from deriving confidence intervals.




\section{Spectral \ucrl}\label{s:learning}

The most interesting aspect of the estimation process illustrated in the previous section is that it can be applied when samples are collected using any policy $\pi$ in the set $\calP$. As a result, it can be integrated into any exploration-exploitation strategy where the policy changes over time in the attempt of minimizing the regret.

\begin{algorithm}[h]
\begin{small}
\setstretch{1.1}
\begin{algorithmic}
\STATE \textbf{Input:} Confidence $\delta'$
\STATE \textbf{Variables:}
\STATE $\quad$ Number of samples $N^{(k)}(l)$
\STATE $\quad$ Estimated observation, reward, and transition models $\wh{f}^{(k)}_O$, $\wh{f}^{(k)}_R$, $\wh{f}^{(k)}_T$
\STATE \textbf{Initialize:} $t=1$, initial state $x_1$, $\delta = \delta'/N^6$, $k=1$
\WHILE{$t < N$}
\STATE Compute the estimated POMDP $\wh{M}^{(k)}$ with the Alg.~\ref{alg:spectral.pomdp} using $N^{(k)}(l)$ samples per action
\STATE Compute the set of admissible POMDPs $\M^{(k)}$ using bounds in Thm.~\ref{thm:estimates}
\STATE Compute the optimistic policy
$\wt{\pi}^{(k)} = \arg\max\limits_{\pi\in\calP}\max\limits_{M\in\M^{(k)}} \eta(\pi; M)$
\STATE Set $v^{(k)}(l) = 0$ for all actions $l\in[A]$
\WHILE{$\forall l\in[A], v^{(k)}(l) < 2N^{(k)}(l)$}
\STATE Execute $a_t \sim f_{\wt{\pi}^{(k)}}(\cdot|\vec{y}_t)$
\STATE Obtain reward $\vec{r}_t$, observe next observation $\vec{y}_{t+1}$, and set $t=t+1$
\ENDWHILE
\STATE Store $N^{(k+1)}(l) = \max_{k'\leq k} v^{(k')}(l)$ samples for each action $l\in[A]$
\STATE Set $k=k+1$
\ENDWHILE
\end{algorithmic}
\caption{The \smucrl algorithm.}
\label{alg:sm.ucrl}
\end{small}
\end{algorithm}

\paragraph{The algorithm.} The \smucrl algorithm illustrated in Alg.~\ref{alg:sm.ucrl} is the result of the integration of the spectral method into a structure similar to \ucrl~\citep{jaksch2010near-optimal} designed to optimize the exploration-exploitation trade-off. The learning process is split into episodes of increasing length. At the beginning of each episode $k>1$ (the first episode is used to initialize the variables), an estimated POMDP $\wh{M}^{(k)} = (X, A, Y, R, \wh{f}_T^{(k)}, \wh{f}_R^{(k)}, \wh{f}_O^{(k)})$ is computed using the spectral method of Alg.~\ref{alg:spectral.pomdp}.

Given the estimated POMDP $\wh{M}^{(k)}$ and the result of Thm.~\ref{thm:estimates}, we construct the set $\M^{(k)}$ of \textit{admissible} POMDPs $\wt{M} = \langle \X, \A, \Y, \R, \wt{f}_T, \wt{f}_R, \wt{f}_O\rangle$ whose transition, reward, and observation models belong to the confidence intervals and compute the optimal policy with respect to optimistic model as follows
\begin{align}\label{eq:optimistic.policy}
\wt{\pi}^{(k)} = \arg\max_{\pi\in\calP}\max_{M\in\M^{(k)}} \eta(\pi; M).
\end{align}
The choice of using the optimistic POMDP guarantees the $\wt{\pi}^{(k)}$ explores more often actions corresponding to large confidence intervals, thus contributing to improve the estimates over time. After computing the optimistic policy, $\wt{\pi}^{(k)}$ is executed until the number of samples for one action is doubled, i.e., $v^{(k)}(l) \geq 2N^{(k)}(l)$. This stopping criterion avoids switching policies too often and it guarantees that when an epoch is terminated, enough samples are collected to compute a new (better) policy. This process is then repeated over epochs and we expect the optimistic policy to get progressively closer to the best policy $\pi^+\in\calP$ as the estimates of the POMDP get more and more accurate.


\paragraph{Regret analysis.}
We now study the regret \smucrl w.r.t.\ the best policy in $\calP$. Given an horizon of $\TN$ steps, the regret is defined as
\begin{align}\label{eq:regret}
\text{Reg}_\TN = \TN\eta^+ - \sum_{t=1}^\TN r_t,
\end{align}
where $r_t$ is the random reward obtained at time $t$ over the states traversed by the policies performed over epochs on the actual POMDP.
To restate, the complexity of learning in a POMDP $M$ is partially determined by its diameter, defined as
\begin{align}\label{eq:diameter}
D:=\max_{x,x'\in\X,a,a'\in\A}\min_{\pi\in\calP}\mathbb{E}\big[\tau(x',a'|x,a; \pi)\big],
\end{align}
which corresponds to the expected passing time from a state $x$ to a state $x'$ starting with action $a$ and terminating with action $a'$ and following the most effective memoryless policy $\pi\in\calP$. Now for result we have the following theorem.


\begin{theorem}[Regret Bound]\label{thm:regret}
Consider a POMDP $M$ with $X$ states, $A$ actions, $Y$ observations, $R$ rewards, characterized by a diameter $D$.
If \smucrl is run over $N$ steps and the confidence intervals of Thm.~\ref{thm:estimates} are used with $\delta = \delta'/N^{6}$ in constructing the plausible POMDPs $\wt{\mathcal M}$, then the total regret
%
\begin{align}\label{eq:regret.bound1}
\text{Reg}_N\leq C_1r_{\max}DX^{3/2}\sqrt{AYN\log (N/\delta')}
\end{align}
with probability $1-\delta'$, where $C_1$ is numerical constants

\end{theorem}



\section{Experiments}\label{s:experiments}
In the following section, we show how $\smucrl$ algorithm outperforms other well-known methods in both synthetic environment and simple computer game.
\subsection{Synthetic Environment}
\begin{figure}[h]
{\vspace*{0cm}
\label{fig:learning}\begin{minipage}{0.55\textwidth}
\begin{center}
\begin{psfrags}
\psfrag{episode}[][1]{\tiny Epochs}
\psfrag{average deviation from true parameter}[][1]{\tiny Average Deviation from True Parameter}
\psfrag{Transition Tensor}[][1]{ \tiny ~~~~~~~  Transition Tensor}
\psfrag{Observation Matrix}[][1]{~~~~~~~~\tiny Observation Matrix}
\psfrag{Reward Tensor}[][1]{\tiny ~~~~ ~~~ Reward Tensor}
\psfrag{Random Policy}[][1]{\tiny Random Policy}
\hspace*{1cm}
\includegraphics[width=7cm,trim={5cm 0cm 0 0},clip]{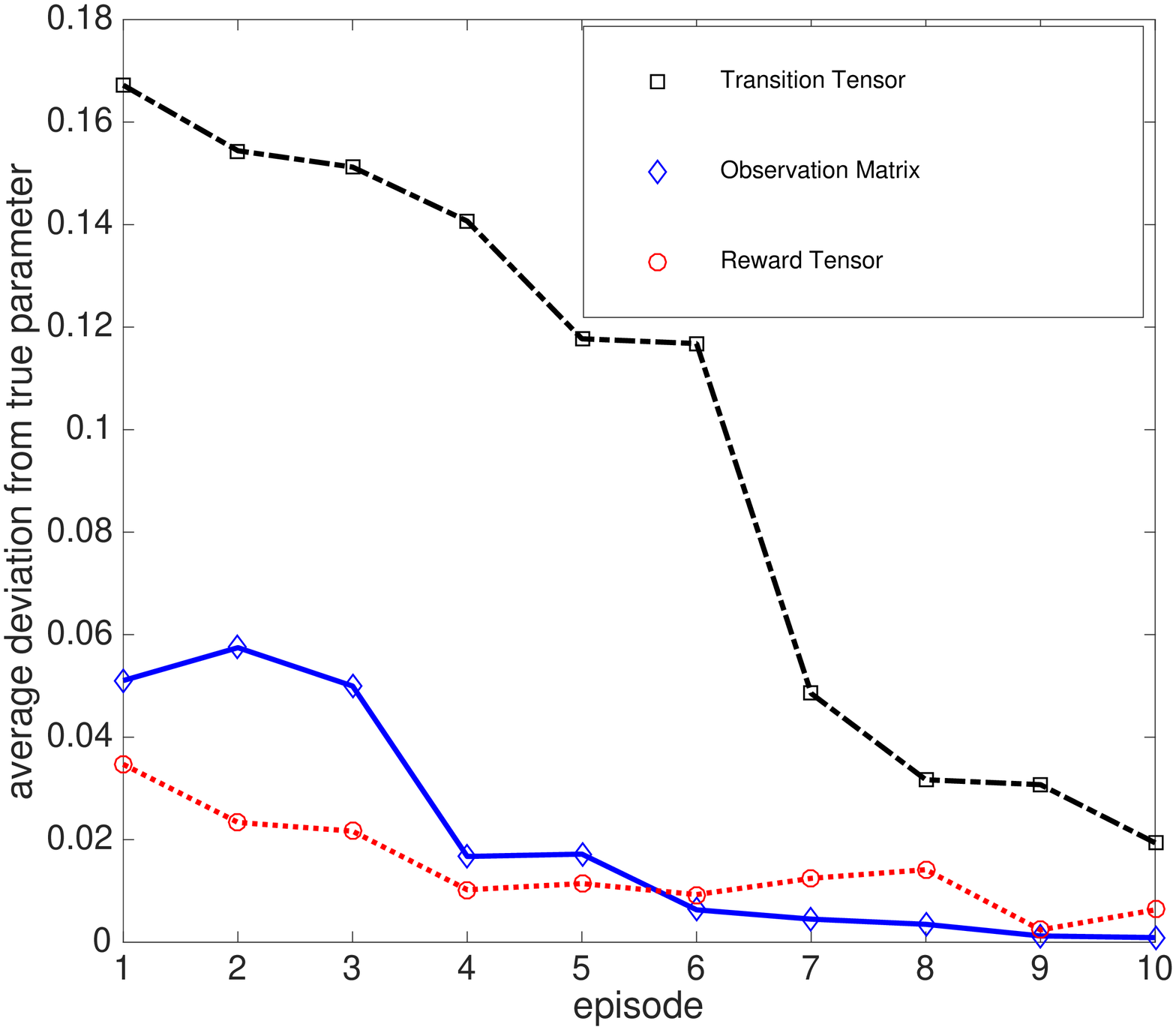}\\
\vspace*{.5cm}
\hspace{-2cm}\small $a)$Learning by Spectral Method
\end{psfrags}
\end{center}
\end{minipage}
\label{fig:performance}
\hspace*{1cm}
\begin{minipage}{0.55\textwidth}
\begin{psfrags}
\psfrag{Average Reward}[][1]{  \tiny Average Reward}
\psfrag{Number of Trials}[][1]{  \tiny Number of Trials}
\psfrag{SM-UCRL-POMDP}[][1]{ {\tiny ~~~~~~  SM-UCRL-POMDP}}
\psfrag{UCRL-MDP}[][1]{ \tiny ~~ UCRL-MDP}
\psfrag{Q-learning}[][1]{  \tiny ~~ Q-Learning}
\psfrag{Random Policy}[][1]{\tiny ~~~~  Random Policy}
\hspace{-1cm}
\includegraphics[width=5.5cm,trim={17cm 0 0cm 1cm},clip]{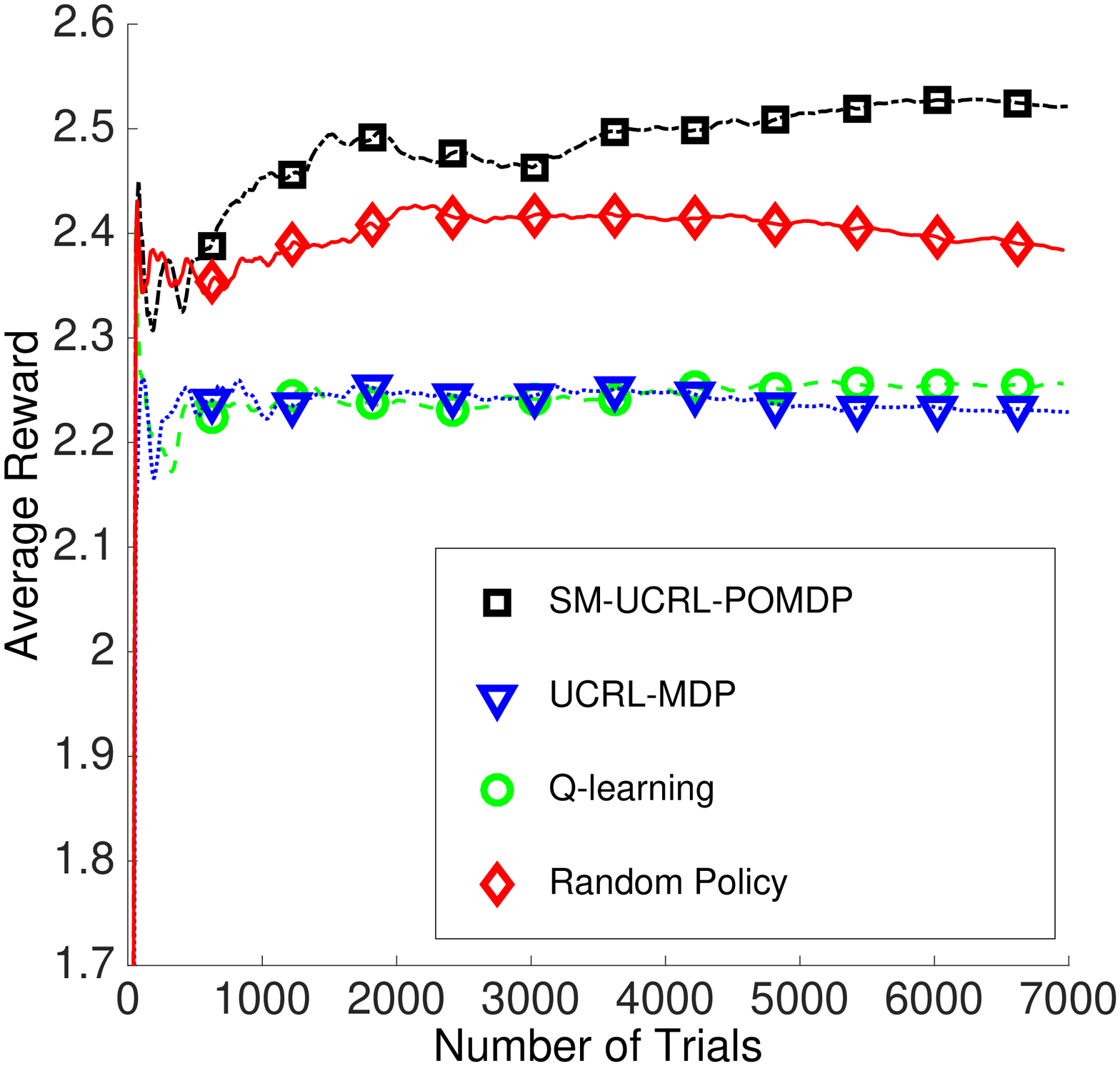}\\
~~~~~~~~~~~~\small            $ ~$ $~ b)$  Regret Performance
\end{psfrags}
\end{minipage}} 
\caption{(a)Spectral estimation error Eqs.~[\ref{eq:obs.bound},\ref{eq:transition.bound}].(b) Average reward Comparison.}
\end{figure}

In this subsection, we illustrate the performance of our method on a simple synthetic environment which follows a POMDP structure with $X=2$, $Y=4$, $A=2$, $R=4$, and $r_{max}=4$.  We find that spectral learning method quickly learn model parameters Fig.~[2]. Estimation of the transition tensor $T$ takes more effort compared to estimation of observation matrix $O$ and reward matrix $R$ due to the fact that the transition tensor is estimated given estimated $O$ matrix which adds up more error.
For planning, given POMDP model parameters, we use alternating maximization method to find the memoryless policy. This method, iteratively, alternates between updates of the policy and the stationary distribution which ends up to stationary point of the optimization problem. We find that, in practice, this method converges to a reasonably good solution (\cite{azizzadenesheli2016open} shows the planing is NP-hard in general). The resulting regret bounds are shown in Fig.~[\ref{fig:performance}].  We compare against the following algorithm: (1) baseline random policies which simply selects random actions without looking at the observed data, (2) \ucrl-MDP \cite{auer2009near} which attempts to fit a MDP model to the observed data and runs the \ucrl policy, and (3) Q-Learning \cite{watkins1992q} which is a model-free method that updates policy based on the Q-function. We find that our method converges much faster. In addition, we show that it converges to a much better policy (stochastic). Note that the MDP-based policies \ucrl-MDP and Q-Learning perform very poorly, and are even worse than a random policy and are too far from $\smucrl$ policy. This is due to model misspecification and dealing with larger state space.
\vspace*{-0.5cm}
\subsection{Simple Atari Game}
In the following, we provide empirical guarantees for $\smucrl$ on environment of simple computer Game. In this game, Figs.~[3], the environment is $10\times 10$ grid world and has five sweet \textit{(green)} apple and five poisonous apples \textit{(red)}. The environment uniformly spread out the apples. In addition, each of these apples lasts uniformly for $(1,2)$ time steps and disappear or they get eaten by the agent. In this game, the agent interacts with this environment. We study two settings, $(i)$ the has set of possible actions (\textit{(N,W,S,E)}, $(ii)$ set of actions \textit{(N,NW,W,SW,S,SE,E,NE)}). At each time step the agent chooses an action and deterministically moves one step in that direction. If there is a sweet apple at the new location, the agent will score up by one, and score down by one if it is poisonous one. In this game, at each time step, the agent just partially observes the environment, Fig.~[3,a] one single box above of the agent is visible to her, and Fig.~[3,b] three boxes above of her are observable. The randomness on rewarding process and partial observability bring the notion of hidden structure and pushes the environment more to be POMPDs models rather than MDPs.
\begin{figure}[h]
{
\begin{minipage}{0.55\textwidth}
\begin{center}
\begin{psfrags}
\psfrag{epoch}[][1]{Epochs}
\psfrag{average deviation from true parameter}[][1]{ Average Deviation from True Parameter}
\psfrag{Transition Tensor}[][1]{ ~~~  \footnotesize Transition Tensor}
\psfrag{Observation Matrix}[][1]{~~~~~~~\footnotesize Observation Matrix}
\psfrag{Reward Tensor}[][1]{\footnotesize ~~~~ Reward Tensor}
\psfrag{Random Policy}[][1]{Random Policy}
\vspace*{-1.5cm}
\includegraphics[width=4cm,trim={0cm 0 0 0},clip]{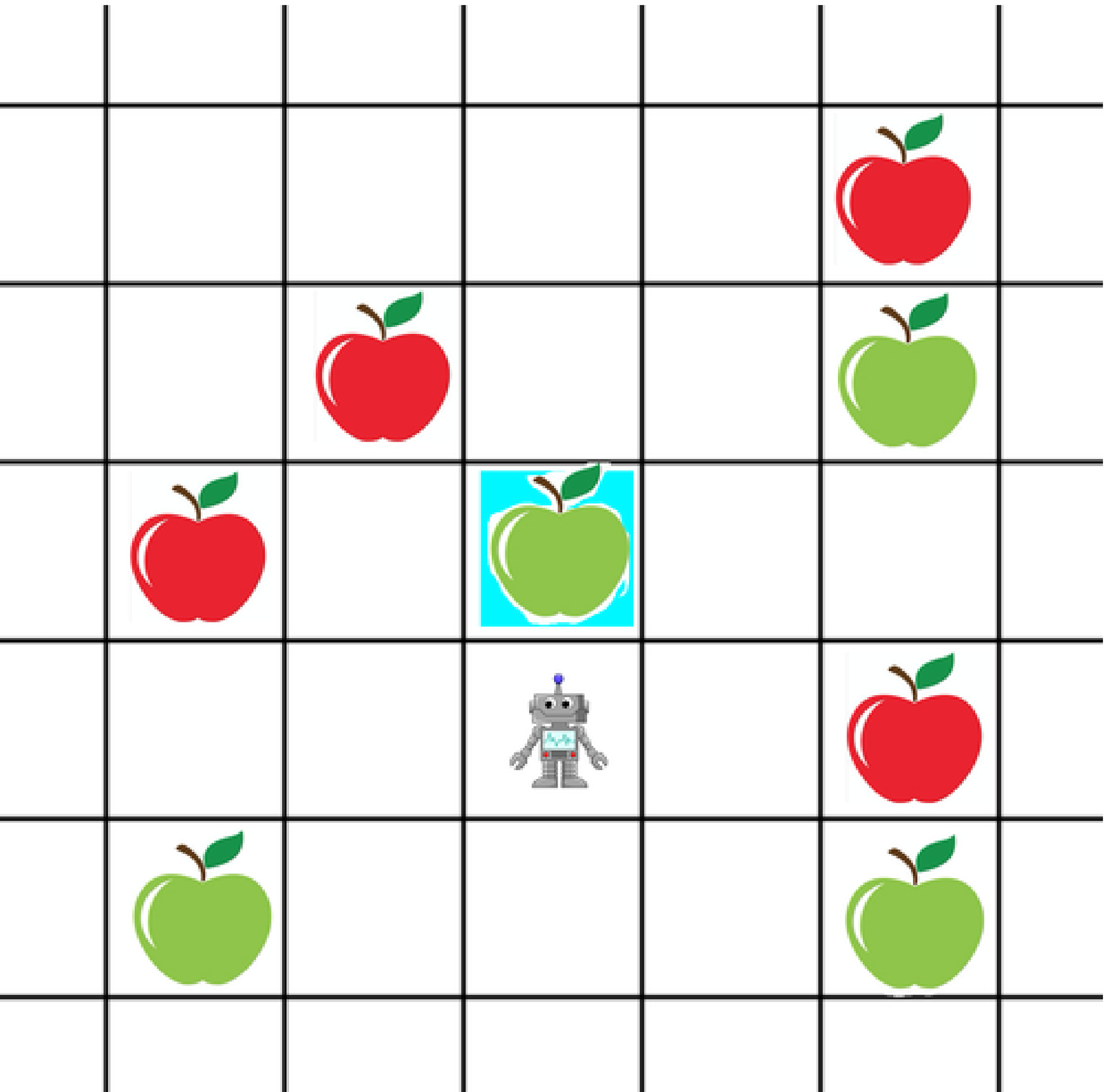}\\
\hspace{-0cm}\small $a)$Single box observable environment
\end{psfrags}
\end{center}
\end{minipage}
\begin{minipage}{0.55\textwidth}
\begin{psfrags}
\psfrag{Average Reward}[][1]{   Average Reward}
\psfrag{Number of Trials}[][1]{   Number of Trials}
\psfrag{SM-UCRL-POMDP}[][1]{ \small ~~~~~~  SM-UCRL-POMDP}
\psfrag{UCRL-MDP}[][1]{ \small ~~ UCRL-MDP}
\psfrag{Q-learning}[][1]{  \small ~~ Q-Learning}
\psfrag{Random Policy}[][1]{\small ~~~~  Random Policy}
\vspace*{-1.5cm}
\includegraphics[width=4cm,trim={0cm 0 0cm 0},clip]{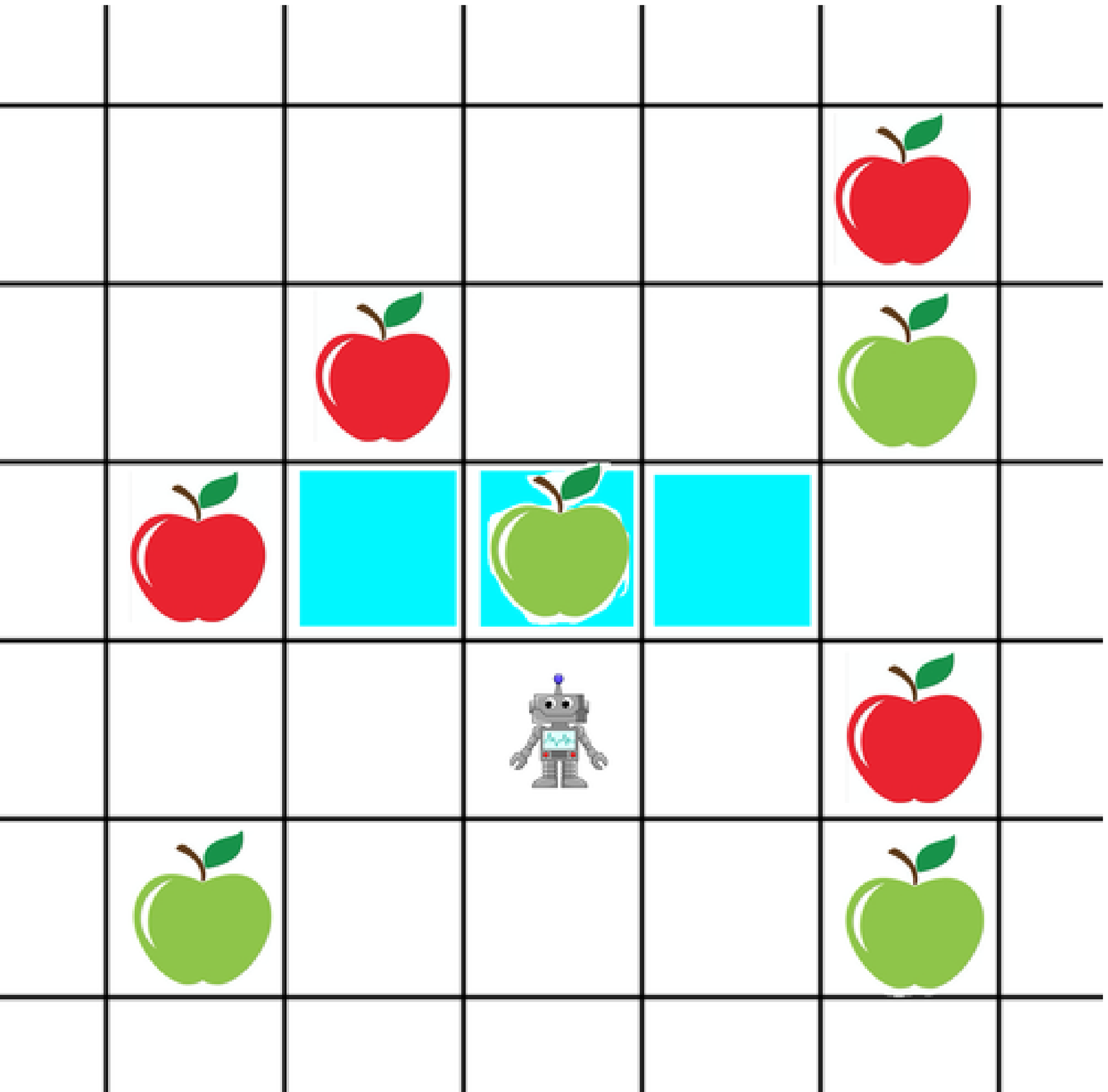}\\
~~~~~~~~~\small $b)$  Triple boxes observable environment
\end{psfrags}
\end{minipage}
} 
\caption{Environment configuration}\label{fig:gamesetups}

\end{figure}

\begin{figure}[!htb]
{
\begin{minipage}{0.32\textwidth}
\begin{center}
\begin{psfrags}
\psfrag{epoch}[][1]{Epochs}
\psfrag{Average Reward}[][1]{   \tiny Average Reward}
\psfrag{time}[][1]{   \tiny Time}
\psfrag{average deviation from true parameter}[][1]{ Average Deviation from True Parameter}
\psfrag{Transition Tensor}[][1]{ ~~~  \footnotesize Transition Tensor}
\psfrag{Observation Matrix}[][1]{~~~~~~~\footnotesize Observation Matrix}
\psfrag{Reward Tensor}[][1]{\footnotesize ~~~~ Reward Tensor}
\psfrag{Random Policy}[][1]{Random Policy}
\includegraphics[width=5.5cm,trim={0cm 0 0 0},clip]{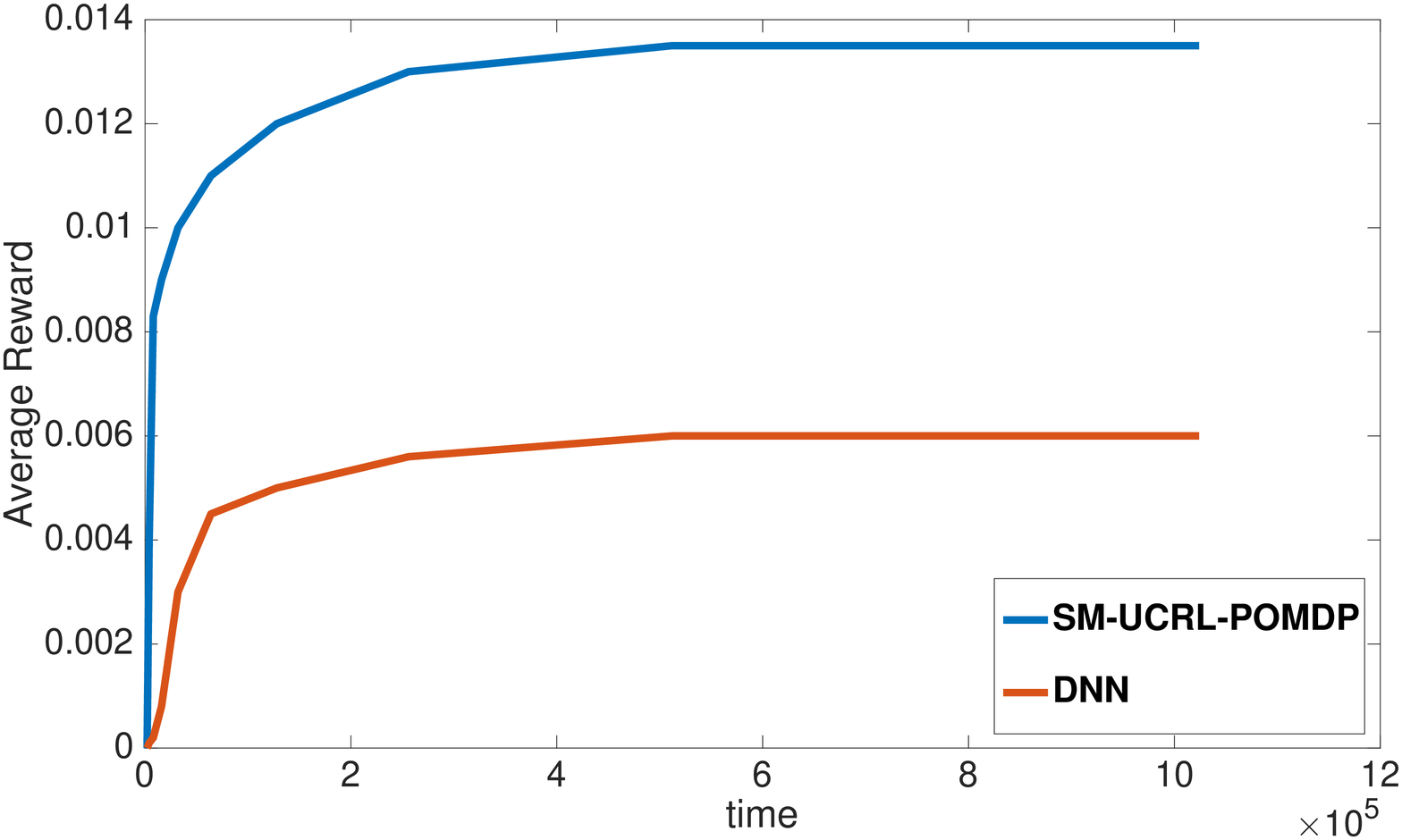}\\
\hspace{-0cm}\small \tiny{$a)$Performances in Single box observable environment.}
\end{psfrags}
\end{center}
\label{fig:single4}
\end{minipage}\hfill
\begin{minipage}{0.32\textwidth}
\begin{psfrags}
\psfrag{Average Reward}[][1]{   \tiny Average Reward}
\psfrag{time}[][1]{   \tiny Time}
\psfrag{Number of Trials}[][1]{   Number of Trials}
\psfrag{UCRL-MDP}[][1]{ \small ~~ UCRL-MDP}
\psfrag{Q-learning}[][1]{  \small ~~ Q-Learning}
\psfrag{Random Policy}[][1]{\small ~~~~  Random Policy}
\hspace{-0cm}
\includegraphics[width=5.5cm,trim={0cm 0 0cm 0},clip]{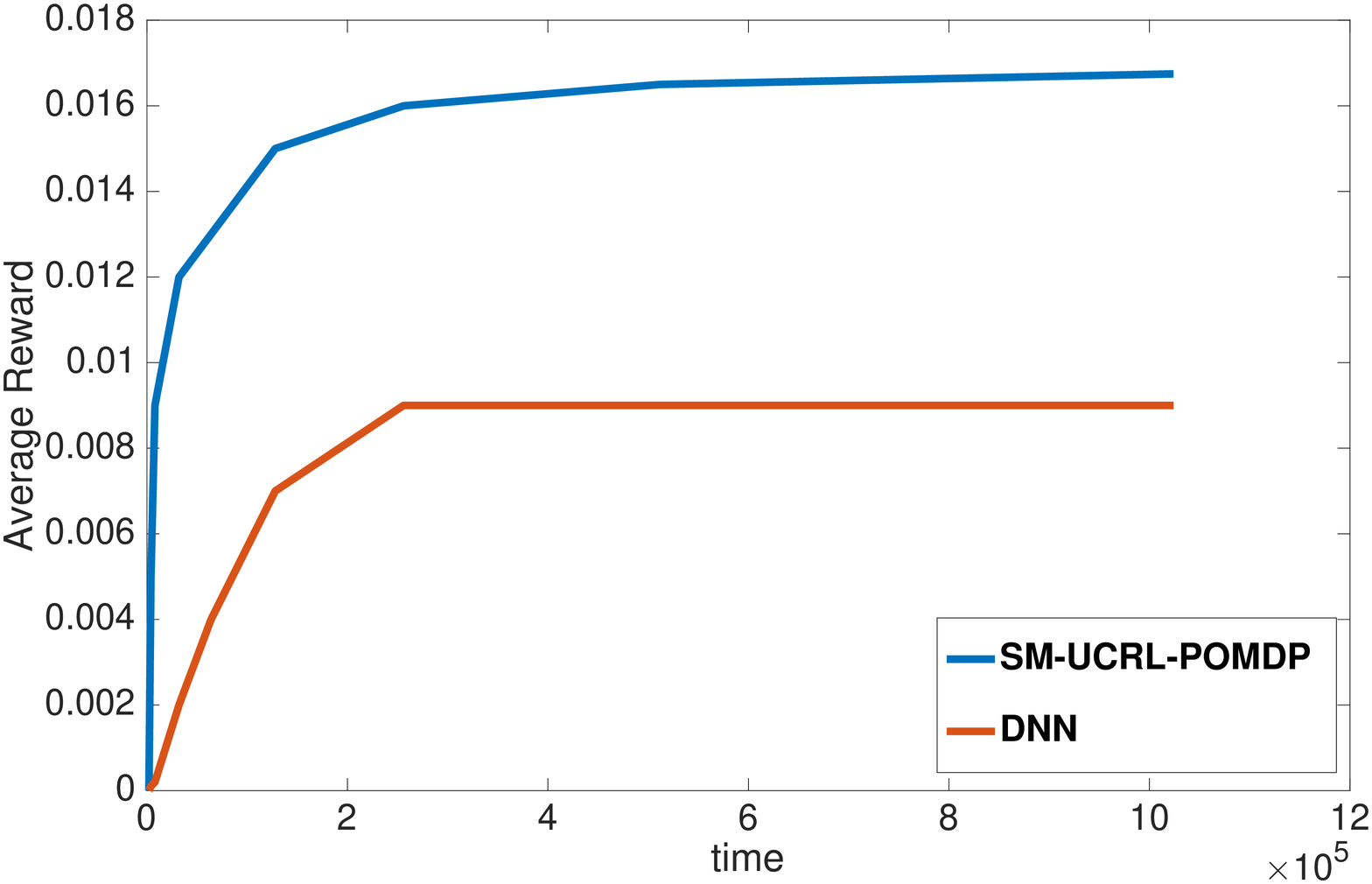}\\
~~~~~~~~~~~~\small           \tiny{ $ ~$ $~ b)$  Performances in Single box observable environment.}
\end{psfrags}
\label{fig:single8}
\end{minipage}\hfill
\begin{minipage}{0.32\textwidth}
\begin{psfrags}
\psfrag{Average Reward}[][1]{   \tiny Average Reward}
\psfrag{time}[][1]{   \tiny Time}
\psfrag{Number of Trials}[][1]{   Number of Trials}
\psfrag{UCRL-MDP}[][1]{ \small ~~ UCRL-MDP}
\psfrag{Q-learning}[][1]{  \small ~~ Q-Learning}
\psfrag{Random Policy}[][1]{\small ~~~~  Random Policy}
\hspace{-0cm}
\includegraphics[width=5.5cm,trim={0cm 0 0cm 0},clip]{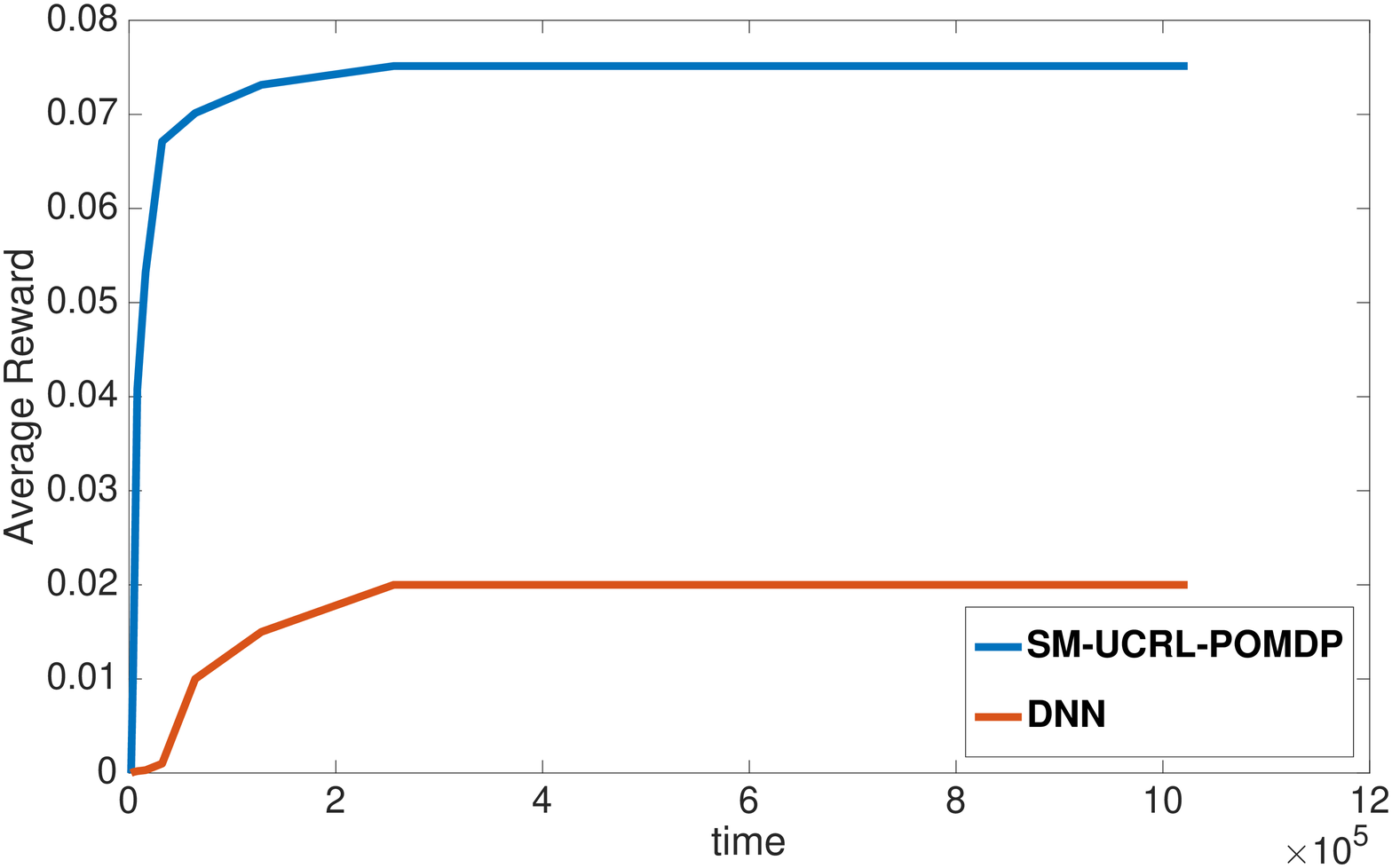}\\
~~~~~~~~~~~~\small           \tiny{ $ ~$ $~ c)$  Performances in triple boxes observable environment.}
\end{psfrags}
\label{fig:triple8}
\end{minipage}} 
\caption{(a) Action set \textit{(N,W,S,E)}, (b,c)Action set \textit{(N,NW,W,SW,S,SE,E,NE)}.}
\end{figure}

For single box observable setting, the observation set has cardinality of 4, \textit{(wall, sweat apple, poisonous apple, nothing)}. We tune $\smucrl$ with \textit{(X=3)}, since it is not known ( we add minimum level of stochasticity to the policy when the policy suggests deterministic mapping to an action). In addition, we apply DQN (Deep Q-Network) \cite{mnih2013playing} with three hidden layers and 10 hidden units of \textit{hyperbolic tangent} activation functions in each hidden layer . For back propagation, we use \textit{RMSProp} method which has been shown to be robust and stable. Figs.~[4] shows the performance of both \smucrl and DQN (DNN) for both case when action set is \textit{(N,W,S,E)} and \textit{(N,NW,W,SW,S,SE,E,NE)}. We show that not only \smucrl captures the environment behavior faster but also reaches to the better long term average reward. We run DQN couple of times and represent the average performance as it is shown in Figs.~[\ref{fig:single4},\ref{fig:single8}]. DQN some times traps in some local minima and it results in bad performance which degrades its average performance.

In other setting, when three boxes are observable Fig.~[3,b], the observation set has cardinality of 64 (four possible observation for each of these three boxes). We tune the $\smucrl$ with \textit{(X=8)} and apply it on this environment. Again, we show that \smucrl outperforms DQN with same structure except 30 hidden units in each hidden layer. During the implementation, we observed that \smucrl does not need to estimate the model parameter very well to get to reasonable policy. It comes up with the stochastic and reasonably good policy even from the beginning. On the other hand, we observed that the policy makes balance between moving upward, downward and makes a good balance between moving rightward and leftward in order to keep the agent away from the walls. It helps the agent to collect more reward and move around the area far from the walls.(link:$https://newport.eecs.uci.edu/anandkumar/pubs/SMvsDQN.flv$)

\vspace*{-0.5cm}


\section{Conclusion}\label{s:conclusions}

We introduced a novel RL algorithm for POMDPs which relies on a spectral method to consistently identify the parameters of the POMDP and an optimistic approach for the solution of the exploration--exploitation problem. For the resulting algorithm we derive confidence intervals on the parameters and a minimax optimal bound for the regret. 

This work opens several interesting directions for future development. POMDP is a special case of the predictive state representation (PSR) model~\cite{littman2001predictive}, which allows representing more sophisticated dynamical systems. Given the spectral method developed in this paper, a natural extension is to apply it to the more general PSR model and integrate it with an exploration--exploitation algorithm to achieve bounded regret. As long as POMDPs are more suitable models for most of real world applications compared to MDP, the further experimental analyses are interesting.

\small
\bibliographystyle{apalike}
\bibliography{NipsWorkshop}

\end{document}